\newif\ifarxiv
\newif\ifralfinal
\ralfinalfalse \arxivtrue

\ifarxiv
\documentclass[letterpaper, 10 pt, conference]{ieeeconf}  %
\fi

\IEEEoverridecommandlockouts                              %
\usepackage{arydshln}
\usepackage{graphics} %
\usepackage{epsfig} %
\usepackage{mathptmx} %
\usepackage{times} %
\usepackage{amsmath} %
\usepackage{amssymb}  %
\usepackage{booktabs}
\usepackage{amsmath}
\usepackage{hyperref}
\usepackage{multirow}
\usepackage{microtype}

\DeclareMathOperator*{\argmax}{arg\,max}
\begin{document}

\title{
\ifarxiv\LARGE \bf\fi
Boosting Performance of a Baseline Visual Place Recognition Technique by Predicting the Maximally Complementary Technique
}

\author{Connor Malone \qquad Stephen Hausler \qquad Tobias Fischer \qquad Michael Milford%
\thanks{All authors are with the QUT Centre for Robotics, School of Electrical Engineering and Robotics, QUT, Brisbane, Australia.}
\thanks{This research was partially supported by funding from an Amazon Research Award, grant 81418190, from ARC Laureate Fellowship FL210100156 to MM, and by funding from Intel’s Neuromorphic Computing Lab to TF and MM. The authors acknowledge continued support from the Queensland University of Technology (QUT) through the Centre for Robotics.
        }%
}

\maketitle
\begin{abstract}
One recent promising approach to the Visual Place Recognition (VPR) problem has been to fuse the place recognition estimates of multiple complementary VPR techniques using methods such as SRAL and multi-process fusion. These approaches come with a substantial practical limitation: they require all potential VPR methods to be brute-force run before they are selectively fused. The obvious solution to this limitation is to predict the viable subset of methods ahead of time, but this is challenging because it requires a predictive signal within the imagery itself that is indicative of high performance methods. Here we propose an alternative approach that instead starts with a \textit{known} single base VPR technique, and learns to predict the most complementary \textit{additional} VPR technique to fuse with it, that results in the largest improvement in performance. The key innovation here is to use a dimensionally reduced difference vector between the query image and the top-retrieved reference image using this baseline technique as the predictive signal of the most complementary additional technique, both during training and inference. We demonstrate that our approach can train a single network to select performant, complementary technique pairs across datasets which span multiple modes of transportation (train, car, walking) as well as to generalise to unseen datasets, outperforming multiple baseline strategies for manually selecting the best technique pairs based on the same training data.
\end{abstract}

\ifralfinal
\begin{IEEEkeywords}
Localization; Visual Place Recognition
\end{IEEEkeywords}
\fi

\section{Introduction}
\label{sec:Introduction}
\ifralfinal
\IEEEPARstart{V}{isual}
\else
Visual
\fi Place Recognition (VPR) is a critical component for robotic and autonomous platforms: providing the ability to recognise previously visited locations solely using image data, which is beneficial for a number of downstream tasks including Simultaneous Localization And Mapping (SLAM) and navigation \cite{lowry2015visual, masone2021survey, zhang2021visual, Garg2021}. VPR is challenging due to the range of appearance and viewpoint changes that can arise when navigating. Appearance changes include diverse seasonal changes, such as summer to winter, or illumination changes like day to night cycles.

Because of the wide variety of appearance and viewpoint changes, it is challenging to design a single approach to VPR that is effective across all possible deployment environments \cite{Garg2021}. Two well known examples~\cite{schubert2021makes} supporting this observation are that strong techniques such as NetVLAD \cite{Arandjelovic2018} perform poorly on the Nordland dataset \cite{Sunderhauf2013} and that the Histogram of Oriented Gradients \cite{DN2005} fail on the Berlin Kudamm dataset \cite{Hausler2020a}. More generally, several works recently highlighted that the optimal VPR technique for a given dataset varies significantly~\cite{Zaffar2021,schubert2021makes}.

\newcommand{\scaleFrontPage}{0.95\linewidth} %
\begin{figure}
    \centering
    \begin{tabular}{c} %
    \includegraphics[width=\scaleFrontPage, clip, trim=70px 85px 120px 50px]{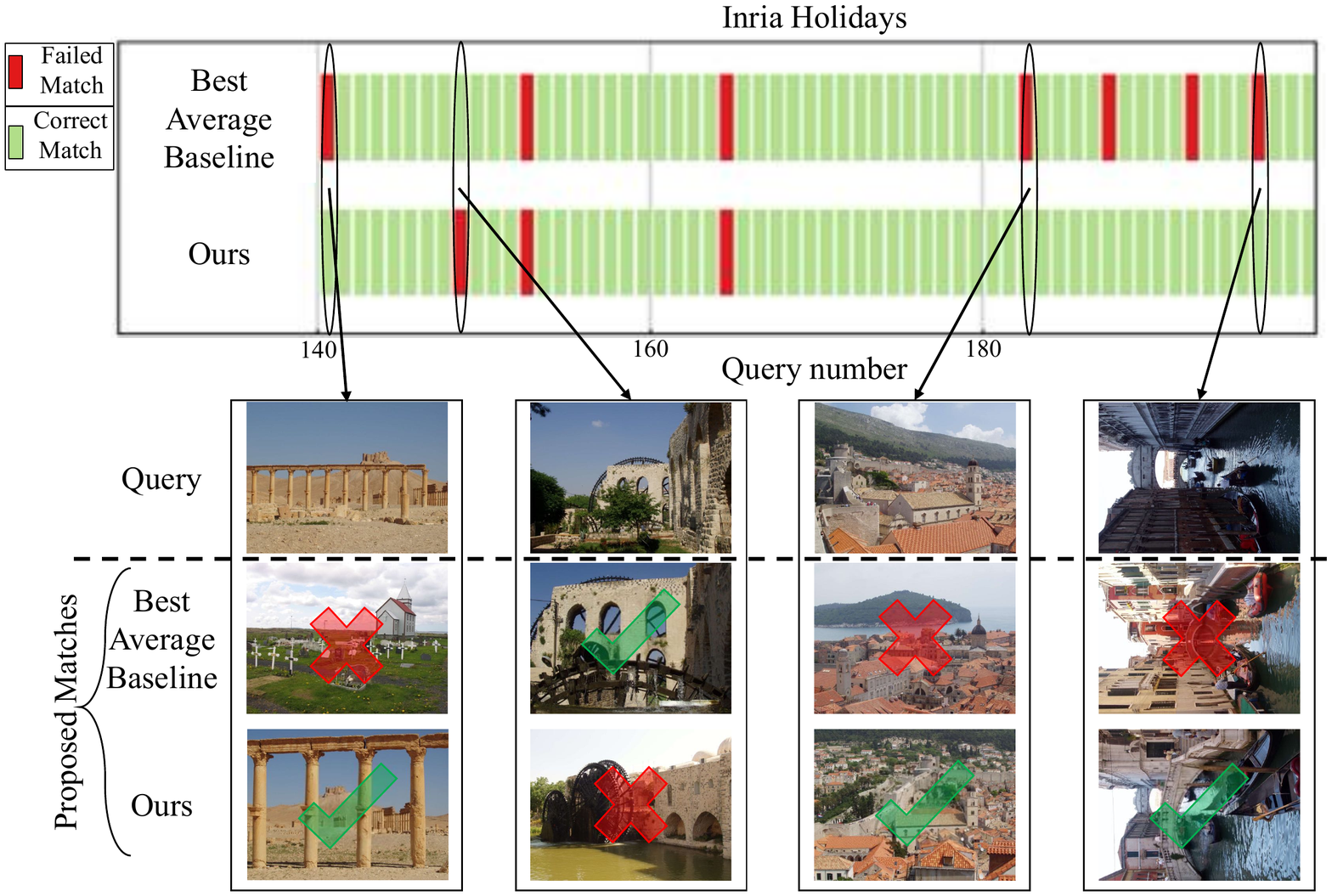}
\end{tabular}
   \caption{Here we demonstrate that our approach for dynamically selecting the maximally complementary VPR technique for a given base technique is able to outperform choosing the average best performing technique in the Inria Holidays dataset. Green bars denote successful place matching and red bars denote failure. We include sample images of the queries and top VPR matches to show where our method outperforms the comparison. We also include a failure case in the second column.}
    \vspace*{-0.3cm}
    \label{fig:frontPage}
\end{figure}

One workaround to this problem is to fuse multiple VPR techniques in tandem, creating a pseudo-multi-sensor fusion with a single sensor \cite{SRAL, Hausler2019}, building on prior multi-sensor fusion approaches \cite{Milford2013a, Zhang, hu2020survey}. By using multiple VPR techniques simultaneously, limitations with one technique (in a specific environment) can be offset with the inclusion of additional techniques. This approach yields performance benefits but has the significant downside of requiring substantive computational capability to run all methods in parallel. Recent work has further shown that certain combinations of techniques are more effective than others on specific datasets -- a phenomenon that has been termed the \emph{complementarity} of different combinations of techniques \cite{SRAL, Hausler2020a, Waheed2021}. 

In this work, we propose a new classification approach that learns to predict the most complementary VPR technique for a known base VPR technique on a frame-by-frame basis. Importantly, this does not require the brute-force running of all techniques beforehand. Our approach first performs VPR using the known base technique to obtain the top matching reference image. The \textit{difference vector} between the query descriptor and retrieved top reference descriptor is used as the input to a multi-label classification network. The network then predicts the best additional VPR technique to fuse with the base technique for VPR using multi-process fusion \cite{Hausler2019}, resulting in an updated place match estimate. We demonstrate that our approach can train a single network to predict complementary technique pairs across datasets captured with varying modes of transport (train, car, walking) and outperform manual selection methods given the same training data. In addition, we show that the same network is capable of generalising to unseen datasets. Our proposed solution enables computationally efficient multi-process fusion to be deployed in various settings for VPR with little training and dynamically chosen technique pairs using a given base technique.

\noindent We make the following specific contributions:

\begin{enumerate}
    \item We introduce a new `anchored' complementarity search concept for VPR, that assumes a baseline technique and searches for a maximally complementary additional technique without requiring brute force running of these technique candidates 
    \item We present a multi-label classification approach that predicts VPR technique utility using a multi-label classification approach, using as input the truncated difference vectors between a query image and image match candidate output by this baseline VPR technique.
    \item We present extensive experimental results demonstrating that the proposed approach outperforms strong baselines across both new parts of familiar environment types but also completely unseen environments.
    \item We provide quantitative analysis of the characteristics of the chosen technique pairs, including their usage distribution within and across different environmental types.
\end{enumerate}
To foster future research in this area, we will provide public source code upon acceptance.

\ifarxiv
\setlength{\topmargin}{-24pt}
\setlength{\headheight}{0pt}
\fi
\section{Related Work}
\label{sec:RelatedWorks}
\subsection{Fusing Multiple Modalities for Improved VPR}
\label{subsec:FusingModalities}Fusing multiple sources of information is a common technique in both SLAM and VPR~\cite{cadena2016past,Garg2021}. Traditionally this has been achieved through the use of multiple physical sensors, like LiDAR, depth cameras, RADAR and Wi-Fi \cite{JA2018, wang2019multi, wei2018multi}. Numerous algorithms have been devised to intelligently fuse these different sensors and produce a more robust localization estimate \cite{Zhang, Milford2013a}. 

Recent work has shown that some of the benefits of multi-sensor fusion can still be attained with a single sensor, by fusing different interpretations of the image data that are obtained by applying different image processing techniques on the same image. SRAL \cite{SRAL} showed that a convex optimization problem can be used to optimize the contribution weights of a set of fused image processing modalities, resulting in improved VPR performance. The optimization was computed for a separate training set for each test dataset. Multi-process fusion \cite{Hausler2019} fuses multiple modalities within a Hidden Markov Model, combining the benefits of both sequences and multiple techniques. Recent works demonstrated that certain combinations of modalities are more effective than others on specific datasets \cite{Hausler2020a, Waheed2021}. However, \cite{Hausler2020a, Waheed2021} calibrate dataset specific combinations and do not generalise to more than one dataset at a time. 

\subsection{Sensor, Technique and Reference Data Selection}
\label{subsec:Consensus}
In \cite{SensorSelect2009}, the sensor selection problem was relaxed from a binary selection to a soft weighted fusion, which enabled the use of convex optimization theorems to approximate the solution to the problem without requiring a brute-force search of all combinations of sensors. In VPR specifically, \cite{Hausler2019a} used a Greedy algorithm to find the optimal set of feature maps to use to improve localization performance, and further work used mutual information to determine the ideal subset of feature maps to use \cite{Maltar2020}. 

\cite{Molloy2021} introduced Bayesian Selective Fusion to intelligently select the optimal reference set of images to use in VPR. Given a set of reference images taken at different times, their algorithm found both an optimal subset of reference times and also weighted the different sets to optimize the localization performance. Another recent work proposes an unsupervised algorithm for automatically selecting the optimal parameters for VPR algorithms \cite{Mount2021}, although only one technique was used at a time. 

\subsection{Place Recognition as a Classification Problem}
\label{subsec:VPRClassification}
Classification is the task of evaluating an image or image region and assigning it a label from a predefined set. Multi-label classification is an extension of this task where labels are not mutually exclusive and an image can have more than one label. We note that we differ from recent work that re-framed place recognition from an image retrieval problem to a classification task to overcome issues such as large scale localisation \cite{berton2022rethinking} and neuromorphic learning \cite{hussaini2022spiking}. In our work, we still frame VPR as an image retrieval problem, but use multi-label classification to select the most complementary technique given a baseline technique.
\section{Proposed Approach}
\label{sec:ProposedApproach}

Our approach aims to maximise VPR performance across varying scenes by fusing a known base technique with the most performant additional technique for a given query using multi-process fusion. This is achieved by training a multi-label classifier to predict the most suitable VPR technique to fuse with the given base technique. The classifier network uses truncated descriptors from the base technique as input features and produces likelihood values for a set of additional VPR techniques indicative of their additive performance potential on top of the base technique. The prediction of the best pairing is taken as the one with the highest likelihood value. We proceed by first providing a brief summary of the multi-process fusion method adopted from previous work and follow with our proposed multi-label classification approach.

\subsection{Multi-Process Fusion}
\label{subsec:MPF}

In VPR, each query image observed by the autonomous platform is compared against a database of $\mathcal{D}$ prior images using a specified feature extraction and matching technique. This produces a $D$ dimensional similarity vector, containing a list of similarity scores between the query image and each database image. A larger score denotes a stronger VPR match.

In a multi-process fusion (MPF) algorithm \cite{Hausler2019}, a variety of different VPR techniques are used simultaneously and a fusion of similarity scores can improve the overall VPR performance of a system. Formally, given a set of techniques $\mathcal{N} = \{1, \dots , N\}$, each technique $n \in \mathcal{N}$ separately computes a similarity vector $\mathbf{s}_{n}$ containing the similarity scores between a query image and a set of database images. Since the set of techniques is arbitrary, the distribution of scores within each technique will not be consistent with the distribution in other techniques. Therefore, a normalization process is used prior to fusing the set of $\mathcal{N}$ techniques, so that each similarity vector has a minimum value of zero and a maximum value of one:
\begin{equation}
    \hat{\mathbf{s}}_{n}(i) = \frac{\mathbf{s}_{n}(i) - \min(\mathbf{s}_{n})}{\max(\mathbf{s}_{n}) - \min(\mathbf{s}_{n})} \quad \forall i ,\quad n \in \mathcal{N}
\end{equation}

\begin{figure}[t]
    \centering
    \includegraphics[width=0.95\columnwidth]{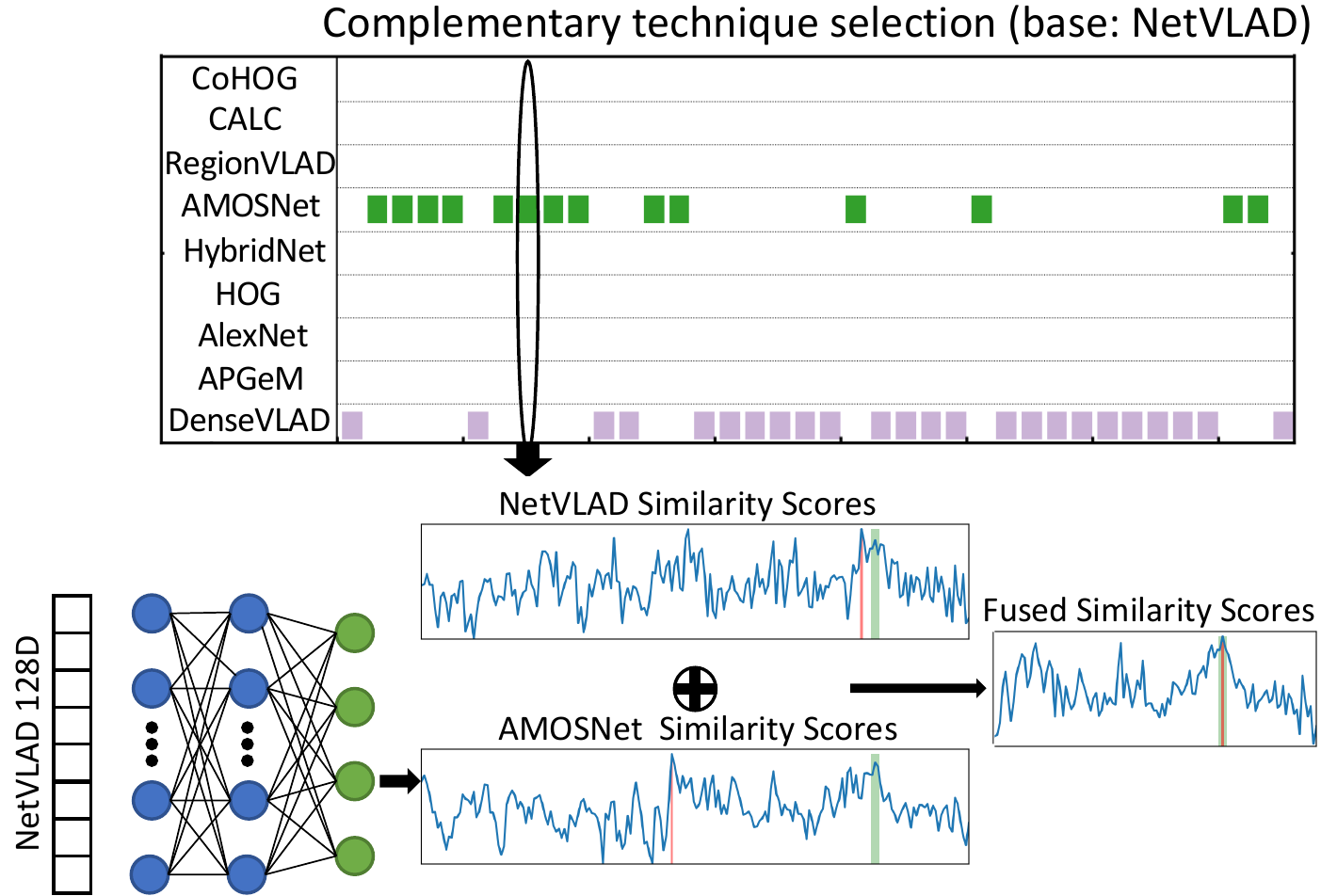} %
    \vspace*{-0.2cm}
    \caption{Given a base technique (NetVLAD), our approach uses a multi-label classification network to determine the most complementary technique pair for VPR using multi-process fusion. \textit{Top:} we demonstrate our approach predicting the most complementary technique pair on a frame-by-frame basis across part of one tested dataset. \textit{Bottom:} for a given query we compute input features using NetVLAD descriptors and use a small classifier network to predict the best technique to pair with NetVLAD. Multi-process fusion is then used to perform VPR using similarity scores for a reference database. In the similarity score plots, the green bar denotes the ground truth and the red bar denotes the top VPR match from the reference database.}
    \label{fig:demo}
    \vspace*{-0.2cm}
\end{figure}

The collection of similarity vectors $\hat{\mathbf{s}}$ are then summed to produce a combined similarity vector:
\begin{equation}
    \mathbf{s}_\text{MPF} = \sum_{n=1}^{\mathcal{N}} \hat{\mathbf{s}}_{n}
\end{equation}
The matching image $X$ is then the image with the maximum score in $\mathbf{s}_\text{MPF}: X=\argmax\ \mathbf{s}_\text{MPF}$ (see Figure \ref{fig:demo}).

While such MPF algorithms have previously been shown to outperform single technique baselines, typically only a sub-selection of techniques will be suited to a particular dataset. The drastic difference in performance between different datasets and even within different sections of the same datasets of particular methods was shown in~\cite{Zaffar2021,Schubert2020}. Therefore, the inclusion of poorly performing techniques (which is the case for standard MPF) can potentially offset any advantages of using multiple methods simultaneously. As an alternative, in the following we describe how the best technique to pair with a given base technique can be dynamically chosen using a multi-label classification approach.

\subsection{Multi-Label Classification}
\label{subsec:MultiLabelClass}
Multi-label classification is a problem where there are multiple correct labels which can all describe a query. This is highly relevant to technique selection in place recognition where multiple separate VPR techniques could all accurately localise a query. In this work, we apply this concept to the problem of finding complementary technique pairs for visual place recognition from the set $\mathcal{N} = \{1, \dots , N\}$ given a base technique, $B \in \mathcal{N}$. For a given query, there are typically multiple pairs of VPR techniques which can localise the query within database $\mathcal{D}$. Therefore we utilise a multi-label classification framework to learn the likelihood that each technique in $\mathcal{N}$ is complementary to the base technique $B$. This is achieved by using descriptors from $B$ as input features.

\subsubsection{Inputs}
\label{subsubsec:ModelInputs}
Our method summarises the difference between query images and the reference database $\mathcal{D}$ using technique $B$ to compute the difference vector $\mathbf{z}_{B, \Delta}$ between a query descriptor $\mathbf{z}_{B, q}$ and the top VPR match descriptor $\mathbf{z}_{B, r}$ as determined by technique $B$.

\begin{equation}
    \mathbf{z}_{B, \Delta} = \mathbf{z}_{B, q} - \mathbf{z}_{B, r}
\end{equation}

To reduce the number of input features used in the network, the difference vector is flattened and truncated. In this work, we use NetVLAD \cite{Arandjelovic2018} as our base technique $B$ and take the first 128 principal components.

At training time we make the adjustment that $\mathbf{z}_{B, r}$ will be the descriptor of the ground truth reference image as computed by $B$, rather than the descriptor of the top VPR match. This removes erroneous training data introduced by incorrect VPR matches from technique $B$.

\subsubsection{Network}
\label{subsubsec:Network}
Given that the chosen input features are already an abstraction of the relevant image data, the network architecture adopted for the following multi-label classifier is a simple multi-layer perceptron model as in Figure \ref{fig:demo}. The model has an input layer with neurons corresponding to the length of $\mathbf{z}_{B, \Delta}$; all hidden layers are given a ReLu activation and dropout; and the output layer has $|\mathcal{N}|$ neurons to represent all techniques which could pair with technique $B$ in the eventual multi-process fusion.

\subsubsection{Training}
\label{subsubsec:Training}
When training the classifier, we utilised the typical methods that are applied in other multi-label classification problems, namely, a sigmoid activation is applied to the networks output to bound the label likelihoods between 0 and 1; and a binary cross entropy ($BCE$) loss is used to train each label output independent of each other:
\begin{equation}
    BCE = -\frac{1}{\mathcal{N}} \sum_{n=1}^{\mathcal{N}}\big(y_n \log\hat{y}_n+(1-y_n)\log(1-\hat{y}_n)\big),
\end{equation}
where $y_n$ is the target for label $n$ and $\hat{y}_n$ is the networks predicted likelihood for label $n$. In this work, label $n$ represents a VPR technique from the set $\mathcal{N}$ and $\hat{y}_n$ is the networks prediction of the complementarity between technique $n$ and technique $B$.

\subsubsection{Prediction}
\label{subsubsec:Prediction}
In the context of visual place recognition, despite training as a multi-label problem, the desired output from the classification network is the single most complementary technique to fuse with technique $B$ for place recognition $\hat{Y}$. This is simply taken as the output neuron/technique from the network with the highest likelihood value.

\begin{equation}
    \hat{Y} = \argmax(\hat{y})
\end{equation}

\section{Experimental Setup}
\label{sec:ExperimentalSetup}
We demonstrate our proposed approach to selecting complementary VPR techniques for a known base technique through a diverse collection of methods and datasets largely provided by the recent VPR-Bench \cite{Zaffar2021}.

\subsection{Datasets}
We use the Gardens Point Walking \cite{SN2015}, Nordland \cite{Sunderhauf2013}, Pittsburgh \cite{Torii2013}, and INRIA Holidays \cite{Jegou2008} benchmark datasets as provided by VPR-Bench \cite{Zaffar2021}. For these datasets, we use the same dataset configurations and ground truth tolerances as used in VPR Bench. In addition, we include traverses from both the Oxford RobotCar \cite{RobotCarDatasetIJRR} and SFU Mountain \cite{sfumountain} datasets. Taken together, these datasets are highly varied, covering a wide range of environments with both traversal and non-traversal (images in the dataset are independent to each other) datasets. A variety of illumination, viewpoint, seasonal and structural changes occur between the reference and query sets of these datasets. For completeness, we provide a brief overview of each dataset:
\subsubsection{Gardens Point Walking}
\label{subsubsec:dsetGPW}
The Gardens Point Walking dataset consists of walking traverses through the Queensland University of Technology, Gardens Point campus. The query and reference sets used in this work include a day and night traverse respectively. As well as strong appearance shift, these images also include viewpoint variations.

\subsubsection{Nordland}
\label{subsubsec:dsetNord}
The Nordland dataset is a common VPR benchmark dataset which covers a 728 kilometer train journey through Norway across multiple seasons. We include subsets from the Winter and Summer traverses as query and reference data in this work.

\subsubsection{Pittsburgh}
\label{subsubsec:dsetPitts}
The Pittsburgh dataset is a large-scale driving dataset which includes full 360 degree viewpoint variations for each query image.

\subsubsection{Inria Holidays}
\label{subsubsec:dsetInria}
Unlike the above datasets, the Inria Holidays dataset is not a continous traversal dataset. It is comprised of images from a wide range of geographically disconnected indoor, outdoor, coastal, densely vegetated and many more types of scenes.

\subsubsection{Oxford RobotCar}
\label{subsubsec:dsetRC}
The Oxford RobotCar set is a long-term localisation dataset which was created by driving a route through central Oxford twice a week for over a year. It covers environment changes such as illumination, seasonal and structural changes. In particular, we adopt a night and day traverse for our query and reference data in this work.

\subsubsection{SFU Mountain}
\label{subsubsec:dsetSFU}
The SFU Mountain dataset consists of more than 8 hours of trail driving captured by a Clearpath Husky robot platform navigating Burnaby Mountain, British Columbia, Canada. It covers a year of changing environmental conditions including sun, rain and snow. For our work we include a snow and sunny traverse as query and reference data.

\subsection{Considered Techniques for Fusion}
For all experiments, we also adopt the same range of hand-crafted and deep-learnt VPR techniques as VPR-Bench. Specifically, we set our base technique $B=$ NetVLAD \cite{Arandjelovic2018} and our further set of techniques $\mathcal{N}$ to include RegionVLAD \cite{Khaliq2020}, CoHOG \cite{Zaffar2020}, HOG \cite{DN2005}, AlexNet, AMOSNet \cite{CZ2017}, HybridNet \cite{CZ2017}, CALC \cite{Merrill}, AP-GeM \cite{Revaud} and DenseVLAD \cite{Torii2018} as defined in Section \ref{subsec:MultiLabelClass}.

\subsection{Multi-Label Datasets}
\label{subsec:Datasets}
In order to train the multi-label classifier for technique selection, we generated multi-hot labels for each of the datasets above, indicative of which technique pairs would localise respective queries. %
For a given query, a technique pair was given a positive label if the top VPR match was within tolerance of the ground truth reference place within $\mathcal{D}$. We limited the technique pairs to those that include the given base technique $B$, NetVLAD.

Each dataset was then split into geographically separated train, validation and testing sets. Since all of the datasets except Inria Holidays consist of sequential frames captured along a route, we maintained geographical separation between the splits by using the first 60\% of queries for training, the next 20\% for validation and the final 20\% for testing purposes. To demonstrate robustness across multiple datasets, we also provide results on a curated dataset where the respective training, testing and validation splits of the Gardens Point Walking, Nordland, Pittsburgh and SFU Mountain datasets were combined. %

To prevent the occurrence of queries without positive labels when training the classification network, we remove queries where none of the technique pairs successfully localised. We emphasise that this does not affect performance trends to favour our method; we simply remove redundant data where none of the technique pairs can possibly be successful and therefore place recognition using these pairs is not possible in any case. Table \ref{tab:dsetStats} shows that the Nordland and Oxford RobotCar datasets are most affected by this curation.

\newcount\columncount
\columncount = 7

\begin{table}[t]
  \footnotesize
  \setlength{\tabcolsep}{2.1pt}
  \renewcommand{\arraystretch}{1.1}
  \centering
  \caption{Number of queries within each dataset}%
  \begin{tabular}{c|cccccc}
  
  \multirow{2}{*}{Dataset}        & Train & Val & Test & Train & Val & Test \\
  & (all) & (all) & (all) & (curated) & (curated) & (curated) \\
  \cline{1-\columncount}
  \cline{1-\columncount}
    SFU Mountain   & 142         & 46        & 50         & 88      & 35            & 47  \\
    GP Walking     & 120         & 39        & 41         & 113     & 33            & 38  \\
    Nordland       & 1656        & 551       & 553        & 707     & 129           & 82  \\
    Pittsburgh     & 600         & 199       & 201        & 585     & 197           & 196 \\
    \cdashline{1-\columncount}
    RobotCar Night & -           & -         & 3876       & -       & -             & 1891 \\
    Inria Holidays & -           & -         & 300        & -       & -             & 280  \\
    \cline{1-\columncount}
    Total          & 2518        & 835       & 5021       & 1493    & 394           & 2534
\end{tabular}%
  \label{tab:dsetStats}%
  \vspace*{-0.2cm}
\end{table}%

\subsection{Baseline Comparisons}
\label{subsubsec:baselines}
For evaluation we compare against two baselines which represent the most likely strategies for manually selecting a single technique pair, inclusive of the base technique $B$ (NetVLAD), to perform visual place recognition across an entire dataset given the same training data. The first baseline we establish for comparison is selecting the NetVLAD technique pair which performs the best on average across the entire training data. For our experiments we found this to be NetVLAD paired with AMOSNet. 

The second baseline which we use for comparison is selecting the best single technique pair, inclusive of the base technique $B$ (NetVLAD), specifically for each dataset based on average performance across the training sets. These were NetVLAD \& Ap-GeM for Gardens Point Walking; NetVLAD \& AMOSNet for Nordland; NetVLAD \& DenseVLAD for Pittsburgh; and NetVLAD \& HybridNet for the SFU Mountain set. Since the RobotCar and Inria Holidays datasets are withheld from training to test as unseen data, we cannot compare them against this second dataset-specific baseline.

\subsection{Metrics}
\label{subsubsec:metrics}
For quantitative results, we use the standard Recall@1 metric which is widely used for evaluating place recognition systems. Recall@1 indicates the percentage of queries which were accurately localised using the top VPR match from the reference database. For this work, the multi-hot labels created for training the classifier network are a query level representation of the Recall@1. Therefore the Recall@1 will be the same as evaluating the accuracy for when the selected technique pair appears as one of the multi-hot targets.

\subsection{Parameter Optimization}
\label{subsec:paramSweep}
During the optimization of network parameters we performed a Bayesian sweep over $batch\:size$, $learning\:rate$, $hidden\:layer\:size$, $number\:of\:hidden\:layers$ and the $dropout$ chance at each hidden layer. To maintain a low model complexity and reduce the chance of over-fitting, the search limited the number of hidden layers to 1-3 and the hidden layer size to 32, 64, 128 or 256. The parameters used for the final model were $batch\:size=8$, $learning\:rate=4.550325e^{-4}$, $hidden\:layer\:size=32$, $number\:of\:hidden\:layers=1$ and $dropout=0.126450$. The network trained for 17 epochs.
\section{Results \& Discussion}
\label{sec:Results}

\subsection{Comparison to Baseline Techniques}
\newcount\columncount
\columncount = 5

\begin{table}[t]
  \footnotesize
  \setlength{\tabcolsep}{2.1pt}
  \renewcommand{\arraystretch}{1.1}
  \centering
  \caption{Recall @ 1 for all Datasets (\%)}%
  \begin{tabular}{c|cccc}
  \multirow{2}{*}{Dataset} & {\multirow{2}{*}{\textbf{NetVLAD}}} & {\textbf{Best average}} & {\textbf{Best dataset-specific}} & {\multirow{2}{*}{\textbf{Ours}}}\\
  & & \textbf{train pair} & \textbf{train pair} & \\
\cline{1-\columncount}
\cline{1-\columncount} 
SFU Mt (Test)            & 57.5  & \textbf{89.4}      & 87.2               & \textbf{89.4}\\
GP Walk (Test)    & 44.7  & 57.9               & 57.9               & \textbf{60.5}\\
Nordland (Test)         & 24.4  & \textbf{61.0}      & \textbf{61.0}      & \textbf{61.0}\\
Pittsburgh (Test)       & 97.5  & 87.2               & \textbf{98.5}      & \textbf{98.5}\\
\cdashline{1-\columncount}
Combined Test           & 57.5  & 78.5      & -               & \textbf{84.8}\\
\cdashline{1-\columncount}
RobotCar Night          & 30.9  & 53.1               & -                  & \textbf{63.1}\\
Inria Holidays          & 81.8  & 90.0               & -                  & \textbf{96.1}\\
\cline{1-\columncount}
\textbf{Mean Recall}    & 56.1  & 73.1               & 74.6               & \textbf{78.1}\\
\end{tabular}%
  \label{tab:oracles}%
  \vspace*{-0.2cm}
\end{table}%

In this section, we present the results for using our multi-label classification network to predict complementary VPR techniques to pair with a given base technique on a frame-by-frame basis. We emphasize that all of the following results have been obtained by training a single network on the combined training set outlined in Section \ref{subsec:Datasets} and testing on both a) a geographically separate test set and b) two datasets from completely unseen environmental types. 

\newcommand{\scaleMaskSets}{0.45\textwidth} %
\begin{figure*}[t]
    \centering
    \begin{tabular}{cc} %

    \includegraphics[width=\scaleMaskSets]{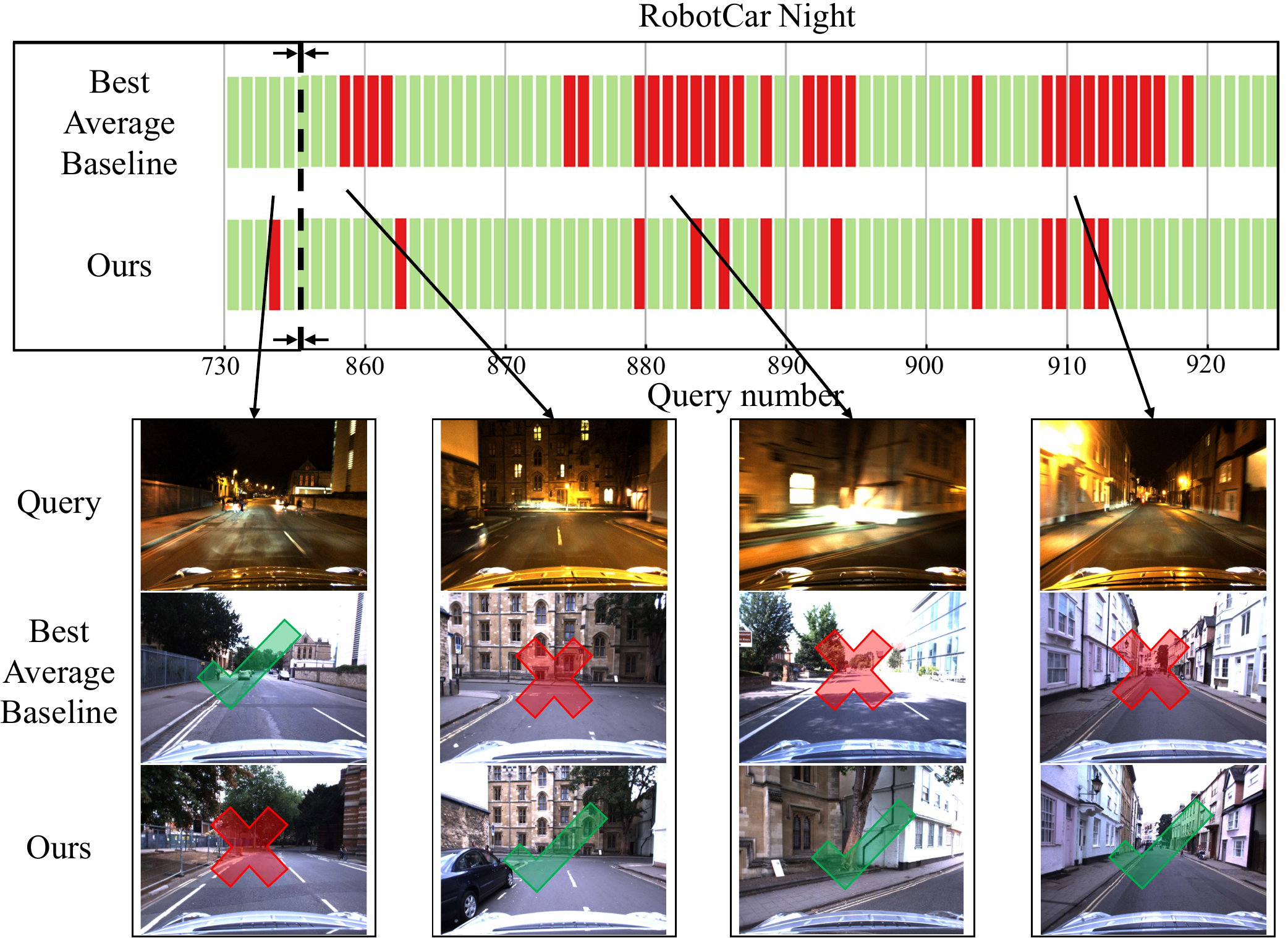} &
    \includegraphics[width=\scaleMaskSets]{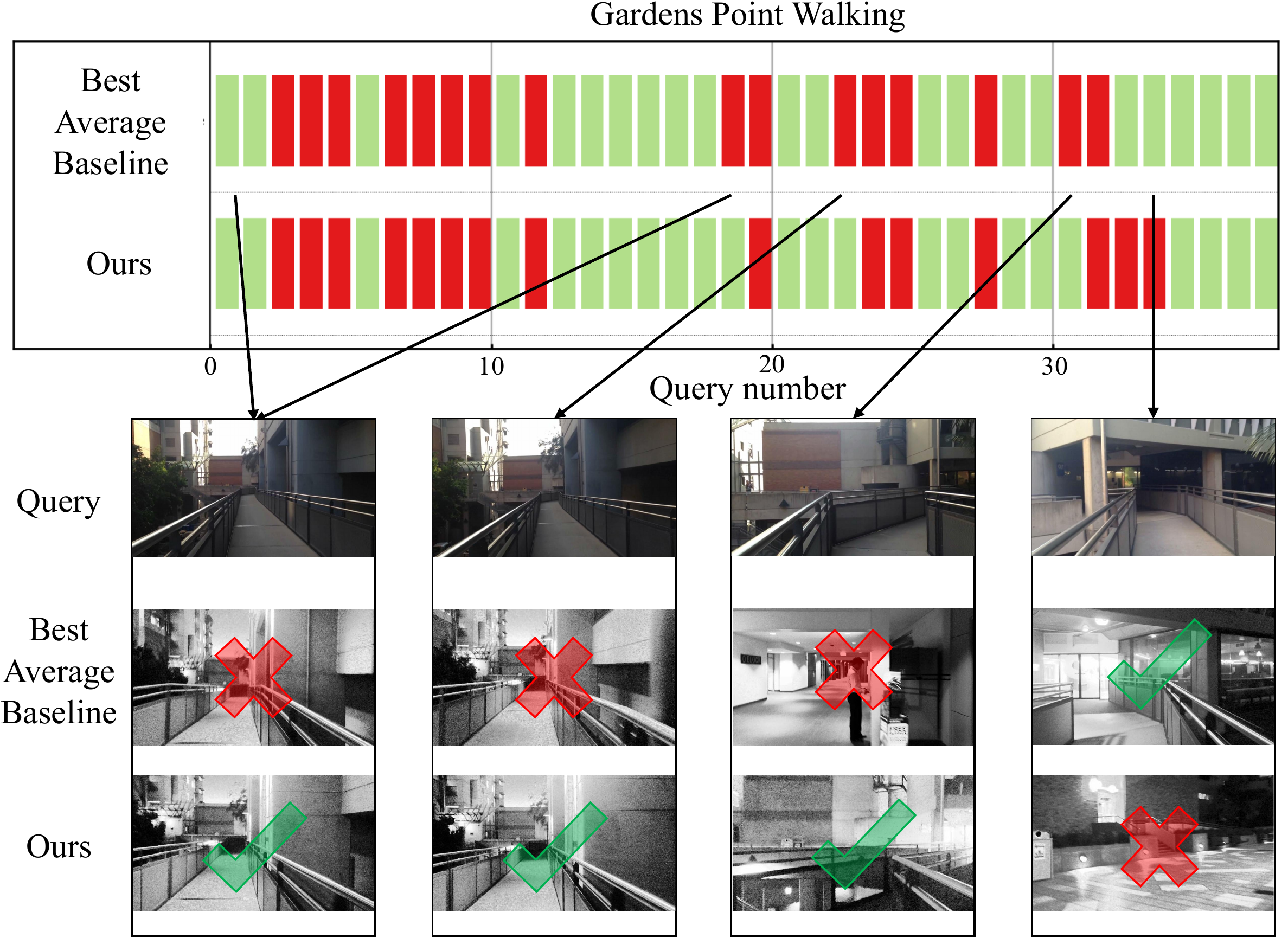} \\
    
\end{tabular}
    \caption{\textbf{Correct/incorrect technique selections per query.} In this figure we show on a query level (each bar denotes an image) where our approach is able to select technique pairs to successfully localise when the baseline did not. We also include some failure examples from our network's technique selection. Green bars denote successful VPR and red bars denote failure. We include two different datasets to demonstrate the extreme changes where we are able to use a single network for complementary technique selection. The datasets will be \textbf{left}: Oxford RobotCar, \textbf{right}: Gardens Point Walking. We include samples from the datasets to show some of the queries and corresponding top VPR matches where these improvements/failures occur.}
    \vspace*{-0.3cm}
    \label{fig:BigTechComp}
\end{figure*}

We start with Table \ref{tab:oracles} which compares the recall of our system against multiple baselines. For the purposes of the `Mean Recall' values, results from the `Combined Test' set are excluded and the `best dataset specific training pair' for the unseen RobotCar and Inria datasets are taken as the `best average training pair' as no information about these sets is available at training time in our experiments. We then provide more detailed discussion on the behaviours of our classifier and performance across the tested datasets as well as query level observations for where our approach does/does not perform better in Figure \ref{fig:BigTechComp}.

\begin{figure}[t]
    \centering
    \includegraphics[width=0.82\columnwidth]{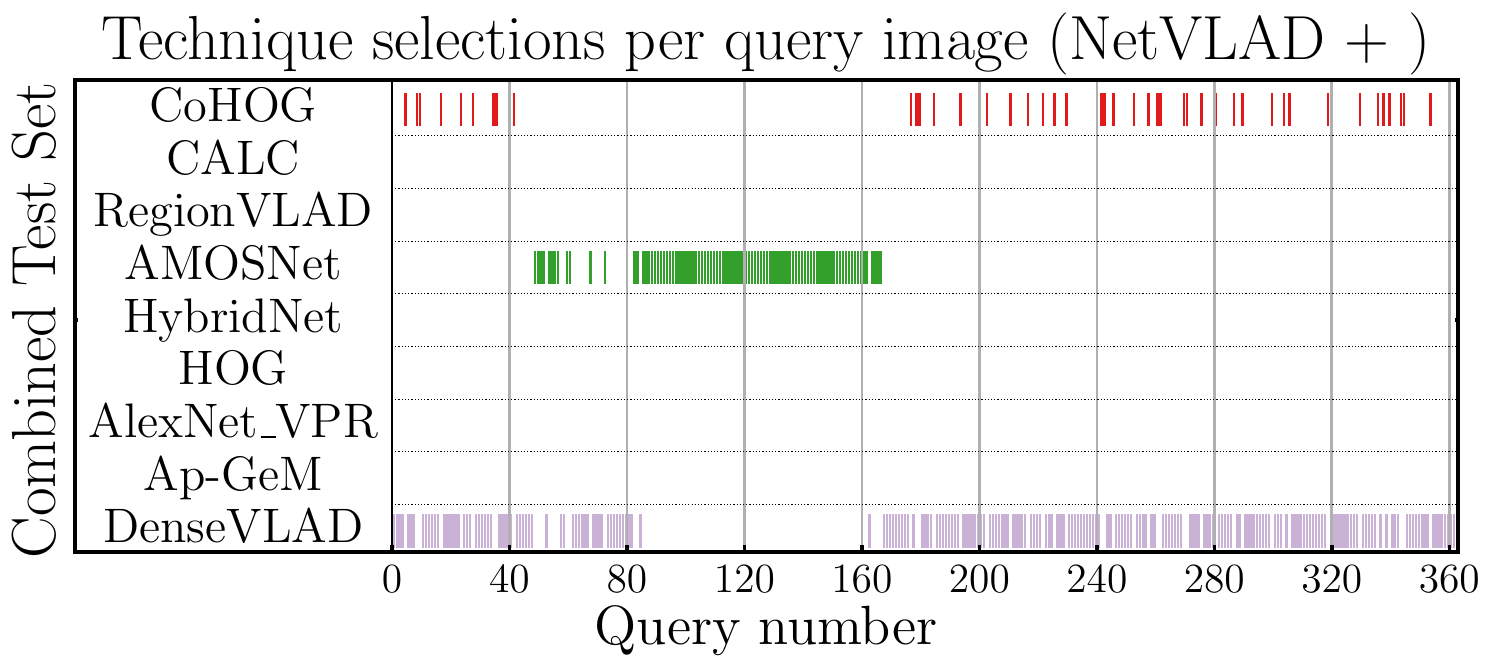}
    \vspace*{-0.2cm}
    \caption{\textbf{Technique selections per query.} Here we show that in the combined test set (made from SFU Mountain, Gardens Point Walking, Nordland and Pittsburgh), our approach is able to use a single network to perform frame-by-frame predictions for the most complementary technique to pair with NetVLAD without any knowledge of the specific dataset queries are from.}
    \label{fig:techSel}
    \vspace*{-0.2cm}
\end{figure}
\begin{figure}[t]
    \centering
    \includegraphics[width=0.75\columnwidth]{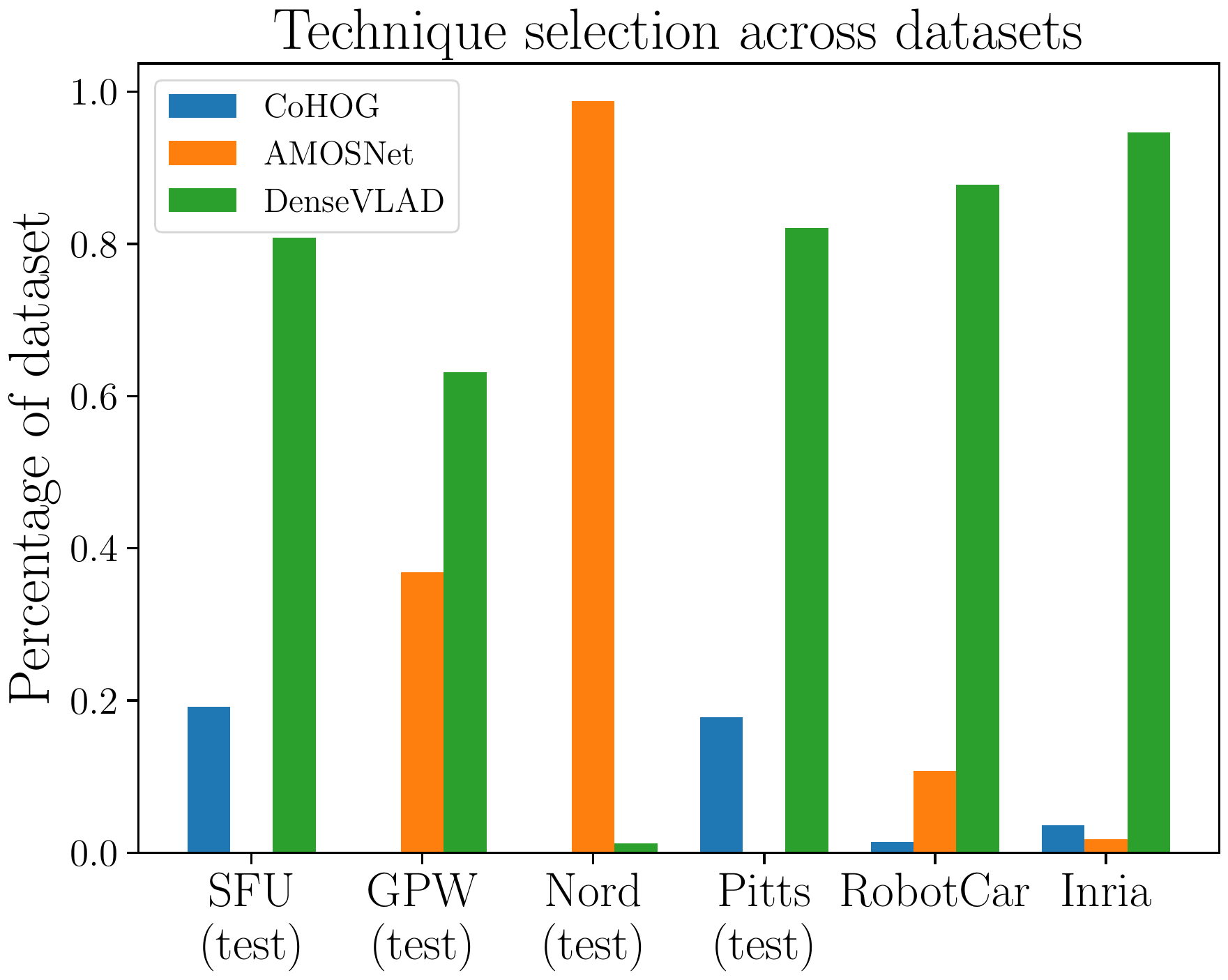} %
    \vspace*{-0.2cm}
    \caption{\textbf{Technique selections per dataset.} In this figure we show the percentage of each dataset where each of the technique pairs were selected using our classification network. We limit this to only include CoHOG, AMOSNet and DenseVLAD as we found our network learned it only needed to make predictions from these three techniques. As seen, there is typically a dominant technique pair which is predicted for individual datasets.}
    \label{fig:TechSelBar}
    \vspace*{-0.2cm}
\end{figure}

Table \ref{tab:oracles} demonstrates that our multi-label classification approach is able to dynamically select techniques which achieve the same or better recall than selecting a static technique pair based on the observed training data. Importantly this table of results shows that neither the best average or dataset specific training pairs (Columns 2 and 3) guaranteed the best pair observed in training generalised to the test sets. However, our classification network is able to at least equal the performance of the best of these statically chosen pairs. Furthermore, in the case of a dataset with a significantly changing environment, such as the `Combined Test' set where the specific dataset is not known, our method provides a significant 6\% improvement in recall by dynamically choosing techniques to pair with NetVLAD. 

This advantage is emphasised by our methods ability to perform in unseen datasets, providing an additional 10\% and 6\% recall in a Day-Night traverse from Oxford RobotCar and the Inria Holidays dataset respectively.

\subsection{Dynamic Technique Selection}
\label{subsec:DynTechSel}
Figure \ref{fig:techSel} illustrates our method's ability to adapt its prediction of the best technique to pair with NetVLAD across significant environment, lighting and viewpoint changes contained within our combined test dataset. Interestingly, our model learnt to only predict three of the nine possible techniques to pair with NetVLAD (CoHOG, AMOSNet and DenseVLAD). We believe this is because the training data shows that there are only 78 of the 1493 training queries where NetVLAD paired with either of these three methods will not be successful whilst other technique pairs are. Therefore, NetVLAD paired with either CoHOG, AMOSNet or DenseVLAD successfully localises in $\approx$95\% of the training data. So whilst we assume the complementarity of different technique pairs to be independent in training, the performance of some technique pairs is likely correlated.

\subsection{Selection Consistency}
\label{subsec:SelCon}
Figures \ref{fig:techSel} \& \ref{fig:TechSelBar} show that within each separated dataset there is typically a particular technique which is primarily selected to pair with NetVLAD. This is not unexpected as environment changes are typically consistent within datasets, such as the seasonal changes in Nordland or the viewpoint changes in Pittsburgh. We recognise that our method is less effective in this scenario, however, when there is significant change within a dataset, such as our combined test set, our method is able to effectively predict the most complementary technique for NetVLAD despite these changes. Table \ref{tab:oracles} quantitatively shows that our method has an improves the mean recall by 3.5\% over the best comparison when dataset specific baselines are used (column 3).

\section{Conclusion and Future Work}
\label{sec:Conclusion}

In this work we propose a multi-label classification approach to selecting the most complementary VPR technique on a frame-by-frame basis to add to a baseline technique in order to maximize place recognition performance using multi-process fusion. The results demonstrated that our approach was effective across datasets with significant environment, illumination, viewpoint and structural changes by providing equal or better recall than manually selected baselines, and that it generalized to two unseen environments.

This work opens up a number of opportunities for future research. One main observation from the experiments was that within individual datasets there was typically one technique pair primarily selected throughout the dataset. While this did result in better performance than the baselines, it was still below the performance of an ``oracle" system (with access to ground truth) which would always choose the best pairing of techniques. This performance gap could potentially be bridged by specifically training input features to aid the multi-label classifier, or through developing loss arguments which encourage a more diverse choice of techniques. 

Finally, while maximizing average performance across an entire dataset is a standard research goal, real-world deployments often prioritise other performance properties like optimizing for worst case performance, minimizing ``time without a good localization match" and other functionally-related properties. Future work could build on the foundation established here to enable any functionally-driven performance goal to be targeted with this approach.

\maxdeadcycles=1000

\IEEEtriggeratref{21}

\bibliographystyle{IEEEtran}
\bibliography{DynMPF}

\end{document}